\documentclass[twoside]{article}

\usepackage{latexsym}
\usepackage{amssymb}
\usepackage{amsmath}
\usepackage{amsthm}
\usepackage{booktabs}
\usepackage{enumitem}
\usepackage{graphicx}
\usepackage{color}
\usepackage{url}
\usepackage{mathtools} 
\usepackage{booktabs} 
\usepackage{amsfonts}
\usepackage{subfigure}
\usepackage{tikz} 
\usepackage{algorithm,algorithmic}
\usepackage{amsthm}
\usepackage{amsfonts}
\usepackage{amssymb}
\usepackage{amsopn}
\usepackage{amsmath}
\usepackage{bbm}

\usepackage[accepted]{aistats2025}
\usepackage[round]{natbib}

\begin{document}

%

%

\twocolumn[

\aistatstitle{The Recurrent Sticky Hierarchical Dirichlet Process Hidden Markov Model}

\aistatsauthor{ Mikołaj Słupiński \And Piotr Lipiński }

\aistatsaddress{ Computational Intelligence Research Group,\\Institute of Computer Science, University of Wrocław,\\ Wrocław, Poland\\ \texttt{\{mikolaj.slupinski,piotr.lipinski\}@cs.uni.wroc.pl}} ]

\begin{abstract}
  The Hierarchical Dirichlet Process Hidden Markov Model
(HDP-HMM) is a natural Bayesian nonparametric extension of the classical Hidden Markov Model for learning from (spatio-)temporal data. A sticky HDP-HMM
has been proposed to strengthen the self-persistence probability in the
HDP-HMM. Then, disentangled sticky HDP-HMM has been proposed to disentangle the strength of the self-persistence prior and transition prior. However, the sticky HDP-HMM assumes that the self-persistence probability is stationary, limiting its expressiveness. Here, we build on previous work on sticky HDP-HMM and disentangled sticky HDP-HMM, developing a more general model: the recurrent sticky HDP-HMM (RS-HDP-HMM). We develop a novel Gibbs sampling strategy for efficient inference in this model. We show that RS-HDP-HMM outperforms disentangled sticky HDP-HMM, sticky HDP-HMM, and HDP-HMM in both synthetic and real data segmentation.
\end{abstract}
\section{INTRODUCTION}
Hidden Markov Models (HMMs) are more and more popular techniques for modeling not only simple time series but also more complex spatio-temporal data. They have been successfully applied to natural language processing \citep{suleiman_use_2017}, speech recognition \citep{yuan_improved_2012}, financial time series analysis \citep{maruotti_hidden_2019}, recognition of objects movement \citep{arslan_semantic_2019, fielding_spatiotemporal_1995}, etc.
\begin{figure}[h]
\subfigure[Sticky HDP-HMM\label{fig:even_ball_seg}]{
  \centering
  \includegraphics[width=0.8\linewidth]{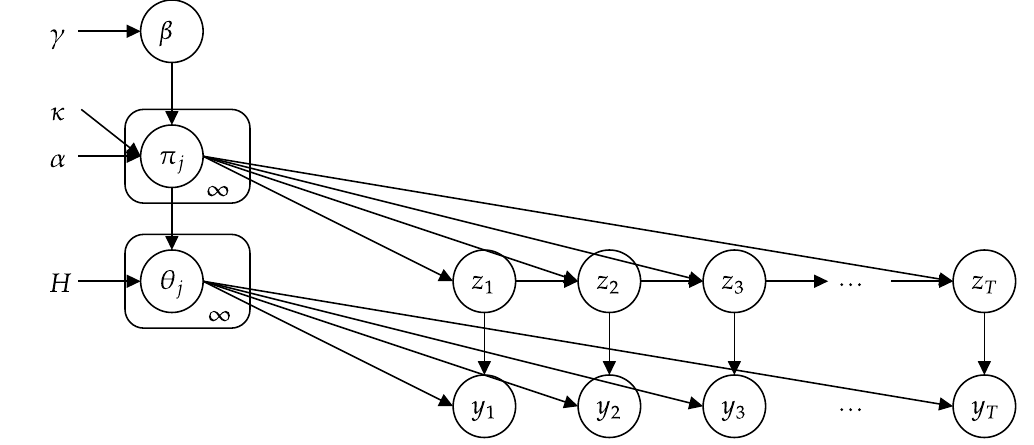}
}%

  \subfigure[Disentangled sticky HDP-HMM\label{fig:even_ball_merged_seg}]{
  \centering
  \includegraphics[width=0.8\linewidth]{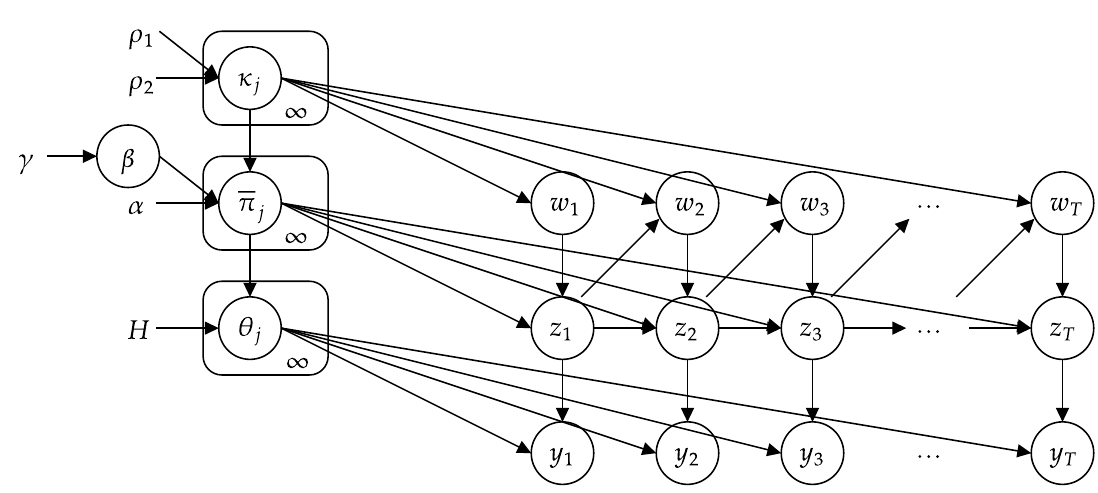}

}
\newline
  \subfigure[Recurrent sticky HDP-HMM\label{fig:pgm:rshdphmm}]{
  \centering
  \includegraphics[width=0.8\linewidth]{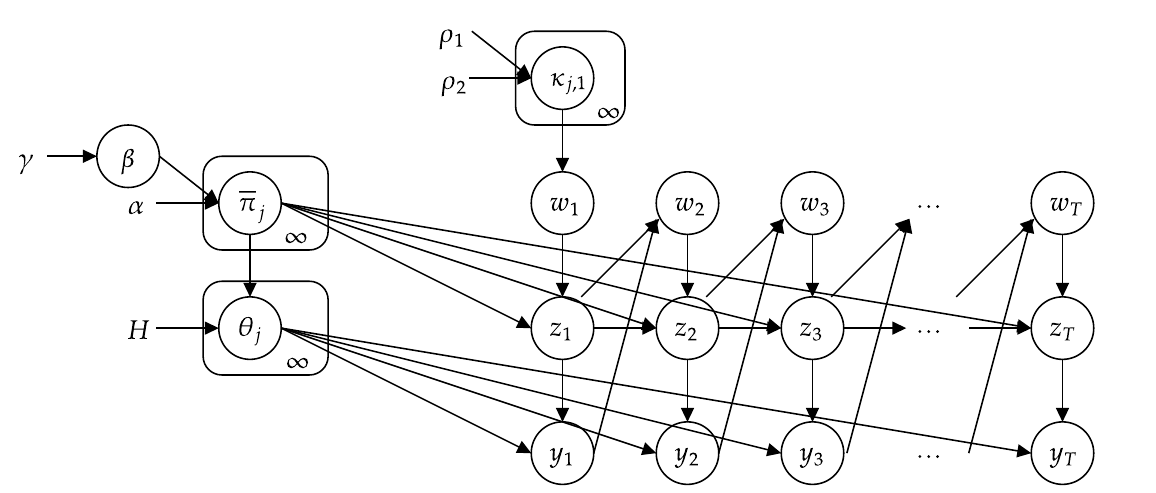}
}%
\caption{Graphical models for sticky HDP-HMM (a) and disentangled sticky HDP-HMM
(b) and recurrent sticky HDP-HMM (c)}%
\label{fig:pgm}

\end{figure}
HMMs posit that the time series values are closely linked to hidden, time-varying states, determining the characteristics of these values. They treat the recorded time series values as observations of a certain random variable, with the current hidden state influencing its probabilistic distribution. Hidden states themselves are modeled as observations of a latent random variable, with its distribution defined by the process dynamics, typically via a transition probability matrix. Learning HMM parameters from recorded data is challenging for complex datasets.

Contemporary methods like Bayesian nonparametric framework \citep{muller_nonparametric_2004} bypass setting the initial number of hidden states, assuming a potentially infinite number. Hierarchical Dirichlet Process Hidden Markov Model (HDP-HMM) \citep{teh_hierarchical_2006}, a key technique, uses the hierarchical Dirichlet process to model the transition matrix. Sticky HDP-HMM (S-HDP-HMM) \citep{fox_bayesian_2009} adds a parameter to control the self-persistence probability of hidden states. Disentangled Sticky HDP-HMM (DS-HDP-HMM) \citep{hutter_disentangled_2021} further refines this by managing the self-persistence prior across hidden states. 

All these methods share a common limitation: they assume stationary state-switching probabilities, which is invalid for spatio-temporal systems where movements depend on position. For instance, whether a car turns is influenced by its current location on the road. Due to their flexibility and interoperability recurrent models are widely used for neural activity analysis \citep{zoltowski_general_2020, osuna-orozco_identification_2023, song_unsupervised_2023, jiang_dynamic_2024, bush_latent_2024} and more recently for vehicle trajectory prediction \citep{wei_navigation_2024}.

In this paper, our main contributions are: a novel approach with recurrent nonparametric modeling, extending Sticky HDP-HMM and Disentangled Sticky HDP-HMM,  two efficient sampling schemes, and comparison of four Bayesian nonparametric models on a sef of benchmarks.
%
\section{BACKGROUND}

\subsection{Hierachical Dirichlet Prior Hidden Markov Models}\label{section:hdp}
The HDP-HMM enables full Bayesian inference of HMMs. The concept is to draw a prior global transition distribution from a Dirichlet process. Then, for each hidden state, a transition distribution is taken from the shared global prior.
We start by introducing some notation for the Dirichlet process (DP).

Given a base distribution $H$ on a parameter space $\Theta$ and a positive concentration parameter $\gamma$, we construct a Dirichlet process $G \sim \mathrm{DP}(\gamma, H)$ (sometimes also denoted by $\mathrm{DP}(\gamma H))$ by the following stick-breaking procedure: let
\begin{equation}
    \beta \sim \operatorname{GEM}(\gamma), \theta_i \stackrel{i i d}{\sim} H, i=1,2, \cdots,
\end{equation}
where $\beta \sim \operatorname{GEM}(\gamma)$ is a random probability mass function (p.m.f.) defined on a countably infinite set as follows:
\begin{equation}
v_i \sim \operatorname{Beta}(1, \gamma), \beta_i=v_i \prod_{l=1}^{i-1}\left(1-v_l\right), i=1,2, \cdots .
\end{equation}
The discrete random measure $G=\sum_i \beta_i \delta_{\theta_i}$ is a sample from $\operatorname{DP}(\gamma H)$, where $\delta_{\theta_i}$ denotes the Dirac measure centered on $\theta_i$.

The HDP-HMM uses the DP to set a prior on the rows of the HMM transition matrix in a situation where the number of latent states may be potentially infinite.  HDP-HMM is defined as
\begin{align}
\text { DP shared global prior: } & \beta \sim \operatorname{GEM}(\gamma), \\
& \theta_j \stackrel{\text { iid }}{\sim} H, j=1,2, \cdots, \\
\text { Transition matrix prior : } & \pi_j \stackrel{i i d}{\sim} \operatorname{DP}(\alpha \beta), j=1,2, \cdots, \\
\text { Latent states : } & z_t \sim \pi_{z_{t-1}}, t=1, \cdots, T, \\
\text { Observations : } & y_t \sim f\left(y \mid \theta_{z_t}\right), t=1, \cdots, T.
\end{align}
Here, $\beta$ and $\left\{\theta_j\right\}_{j=1}^{\infty}$ are specified as in the DP previously described, and then each transition distribution $\pi_j$ for the state $j$ is taken as a random sample from a second DP with the base measure $\beta$ and the concentration parameter $\alpha$. This parameter $\alpha$ determines how close $\pi_j$ is to the global transition distribution $\beta$.
At time $t$, the state of a Markov chain is indicated by $z_t$, and the observation $y_t$ is independently distributed given the latent state $z_t$ and the parameters $\left\{\theta_j\right\}_{j=1}^{\infty}$, with the emission distribution $f(\cdot)$. The sticky HDP-HMM from \citet{fox_bayesian_2009} modifies the transition matrix prior by introducing a point mass distribution with a stickiness parameter $\kappa$ to encourage state persistence. This is done by setting the transition matrix prior to $\pi_j \sim \mathrm{DP}\left(\alpha \beta+\kappa \delta_j\right), j=1,2, \cdots$, where $\delta_j$ is the Dirac measure centered on $j$.

\citet{hutter_disentangled_2021} proposed a new model - Disentangled Sticky HDP-HMM, separating the strength of self-persistence from the similarity of the transition probabilities. The authors modified the transition matrix prior as
\begin{align}
    & \kappa_j \stackrel{\text { iid }}{\sim} \operatorname{beta}\left(\rho_1, \rho_2\right), \\
& \bar{\pi}_j \stackrel{i i d}{\sim} \mathrm{DP}(\alpha \beta), \\
& \pi_j=\kappa_j \delta_j+\left(1-\kappa_j\right) \bar{\pi}_j, j=1,2, \cdots,
\end{align}
where the transition distribution $\pi_j$ is a mixture distribution. A sample from $\pi_j$ has the probability of self-persistence of $\kappa_j$ coming from a point mass distribution at $j$, and has the probability $1-\kappa_j$ coming from $\bar{\pi}_j$, a sample from DP with a base measure $\beta$. We call this new model the disentangled sticky HDP-HMM.

\subsection{Hierachical Dirichlet Prior Hidden Markov Models}\label{section:hdp}
The HDP-HMM enables full Bayesian inference of HMMs. The concept is to draw a prior global transition distribution from a Dirichlet process. Then, for each hidden state, a transition distribution is taken from the shared global prior.
We start by introducing some notation for the Dirichlet process (DP).

Given a base distribution $H$ on a parameter space $\Theta$ and a positive concentration parameter $\gamma$, we construct a Dirichlet process $G \sim \mathrm{DP}(\gamma, H)$ (sometimes also denoted by $\mathrm{DP}(\gamma H))$ by the following stick-breaking procedure: let
\begin{equation}
    \beta \sim \operatorname{GEM}(\gamma), \theta_i \stackrel{i i d}{\sim} H, i=1,2, \cdots,
\end{equation}

where $\beta \sim \operatorname{GEM}(\gamma)$ is a random probability mass function (p.m.f.) defined on a countably infinite set as follows:
\begin{equation}
v_i \sim \operatorname{Beta}(1, \gamma), \beta_i=v_i \prod_{l=1}^{i-1}\left(1-v_l\right), i=1,2, \cdots .
\end{equation}
The discrete random measure $G=\sum_i \beta_i \delta_{\theta_i}$ is a sample from $\operatorname{DP}(\gamma H)$, where $\delta_{\theta_i}$ denotes the Dirac measure centered on $\theta_i$.

The HDP-HMM uses the DP to set a prior on the rows of the HMM transition matrix in a situation where the number of latent states may be potentially infinite.  HDP-HMM is defined as
\begin{align}
\text { DP shared global prior: } & \beta \sim \operatorname{GEM}(\gamma), \\
& \theta_j \stackrel{\text { iid }}{\sim} H, j=1,2, \cdots, \\
\text { Transition matrix prior : } & \pi_j \stackrel{i i d}{\sim} \operatorname{DP}(\alpha \beta), j=1,2, \cdots, \\
\text { Latent states : } & z_t \sim \pi_{z_{t-1}}, t=1, \cdots, T, \\
\text { Observations : } & y_t \sim f\left(y \mid \theta_{z_t}\right), t=1, \cdots, T.
\end{align}
Here, $\beta$ and $\left\{\theta_j\right\}_{j=1}^{\infty}$ are specified as in the DP previously described, and then each transition distribution $\pi_j$ for the state $j$ is taken as a random sample from a second DP with the base measure $\beta$ and the concentration parameter $\alpha$. This parameter $\alpha$ determines how close $\pi_j$ is to the global transition distribution $\beta$.
At time $t$, the state of a Markov chain is denoted by $z_t$, and the observation $y_t$ is independently distributed given the latent state $z_t$ and parameters $\left\{\theta_j\right\}_{j=1}^{\infty}$, with emission distribution $f(\cdot)$. The sticky HDP-HMM from \citet{fox_bayesian_2009} modifies the transition matrix prior by introducing a point mass distribution with a stickiness parameter $\kappa$ to encourage state persistence. This is done by setting the transition matrix prior to $\pi_j \sim \mathrm{DP}\left(\alpha \beta+\kappa \delta_j\right), j=1,2, \cdots$, where $\delta_j$ is the Dirac measure centered on $j$.

\citet{hutter_disentangled_2021} proposed a new model - Disentangled Sticky HDP-HMM, separating the strength of self-persistence from the similarity of the transition probabilities. The authors modified the transition matrix prior as
\begin{align}
    & \kappa_j \stackrel{\text { iid }}{\sim} \operatorname{beta}\left(\rho_1, \rho_2\right), \\
& \bar{\pi}_j \stackrel{i i d}{\sim} \mathrm{DP}(\alpha \beta), \\
& \pi_j=\kappa_j \delta_j+\left(1-\kappa_j\right) \bar{\pi}_j, j=1,2, \cdots,
\end{align}
where the transition distribution $\pi_j$ is a mixture distribution. A sample from $\pi_j$ has the probability of self-persistence of $\kappa_j$ coming from a point mass distribution at $j$, and has the probability $1-\kappa_j$ coming from $\bar{\pi}_j$, a sample from DP with a base measure $\beta$. We call this new model the disentangled sticky HDP-HMM.

\subsection{Limitations of HDP-HMM, sticky HDP-HMM and disentangled sticky HDP-HMM}

The HDP-HMM uses the concentration parameter $\alpha$ to modify the prior strength of the transition matrix or the similarity of the rows of the transition matrix. A high value $\alpha$ implies that the transition probability for each state is close to the global transition distribution $\beta$.

The sticky HDP-HMM introduces a parameter $\kappa$ in comparison to the HDP-HMM. The ratio $\kappa /(\alpha+\kappa)$ determines the average probability of self-persistence or the mean of the diagonal of the transition matrix. Both the similarity of the
rows of the transition matrix and the strength of prior self-persistence are regulated by $\alpha+\kappa$. The beta prior of disentangled sticky HDP-HMM $\left(\rho_1, \rho_2\right)$ has the ability to control both the expectation of self-persistence and the variability of self-persistence. At the same time, $\alpha$ is free to control the variability of the transition probability around the mean transition $\beta$.

All of the models based on HDP-HMM assume that the state transition is stationary. This is especially problematic for spatio-temporal systems, where the states may be dependent on the position. To clarify, our objective is to introduce a dependency of our self-persistence parameter $\kappa$ on the observation $y_t$. This can be achieved by employing logistic regression.

\subsection{Pólya-Gamma augmentation}

One strategy that has been employed in other recurrent models for enabling efficient and quick inference \citep{linderman_bayesian_2017,nassar_tree-structured_2019} involves the utilization of Pólya-Gamma augmentation.

The main result of \citet{polson_bayesian_2013} is that binomial probabilities can be expressed as a combination of Gaussians in terms of a Pólya-Gamma distribution. The fundamental integral identity that lies at the heart of their discovery is that, for $b>0$,
\begin{equation}\label{eq:polya_gamma}
    \frac{\left(e^\psi\right)^a}{\left(1+e^\psi\right)^b}=2^{-b} e^{\lambda \psi} \int_0^{\infty} e^{-\omega \psi^2 / 2} p(\omega) d \omega.
\end{equation}
The value of $\lambda$ is calculated as the difference between $a$ and $b$ divided by two. The conditional distribution of $\omega$ given $\psi$ is a Pólya-Gamma one, which makes it possible to use a Gibbs sampling approach for a variety of binomial models. This requires drawing from a Gaussian distribution for the main parameters and from a Pólya-Gamma distribution for the latent variables.

Suppose the observation of the system at time $t$ follows
\begin{equation}\label{eq:logisticregression}
\begin{gathered}
p\left(w_{t+1} \mid x_t \right)=\operatorname{Bern}\left(\sigma\left(v_t\right)\right)=\frac{\left(e^{v_t}\right)^{w_{t+1}}}{1+e^{v_t}}, \\
v_t=R^T x_t +r,
\end{gathered}    
\end{equation}
where $R \in \mathbb{R}^{d_x}, r \in \mathbb{R}$, $\sigma$ is logistic function, and $\operatorname{Bern}$ denotes Bernoulli distribution.
Then if we introduce the PG auxiliary variables $\eta_{n, t}$, conditioning on $\eta_{1: T}$, \eqref{eq:logisticregression} becomes
$$
\begin{aligned}
p\left(w_t \mid x_t, \eta_{t}\right) & =e^{-\frac{1}{2}\left(\eta_{t} v_{t}-2 \omega_{t} v_{t}\right)} \\
& \propto \mathcal{N}\left(R^T x_t+r \mid \omega_{t} / \eta_{t}, 1 / \eta_{t}\right),
\end{aligned}
$$
where  $\omega_{t}=w_{t}-\frac{1}{2}$.
\section{Related work}

In recent years, significant work has focused on recurrent modeling. \citet{linderman_bayesian_2017} introduced a method for modeling recurrence in recurrent switching linear dynamical systems (rSLDS, also known as augmented SLDS \citep{barber_expectation_2005}), reducing it to a recurrent autoregressive HMM (rARHMM) via PG augmentation, similar to our approach.

Other approaches use neural networks for recurrent connections, including recurrent Hidden Semi Markov Models \citep{dai_recurrent_2016}, Switching Nonlinear Dynamical Systems \citep{dong_collapsed_2020}, and Recurrent Explicit Durations Switching Dynamical Systems \citep{ansari_deep_2021}. Recently, \citet{geadah_parsing_2023} proposed the infinite recurrent switching linear dynamic system (irSLDS). Both methods address recurrence in switching dynamics but differ substantially. Our model incorporates recurrence to influence self-persistence, unlike irSLDS, which integrates it into the state switch probability (see Eq.4).

We use Pólya-Gamma augmentation for efficient inference, whereas in irSLDS this results in intractable inference (see the Appendix of \citep{geadah_parsing_2023}). Additionally, irSLDS uses the variational Laplace-EM algorithm for approximate posterior fitting, while we offer two MCMC sampling algorithms.%


\section{Recurrent sticky HDP-HMM}

One of the problems of models based on HDP-HMM is that they assume stationarity of state-transition. It is especially problematic in a setting where we try to model motion.

Let us imagine that we want to model the car's behavior. We want to make the vehicle more likely to be in the "turn" state, when it is closer to the obstacle.

To achieve position dependence we condition self-persistence on the previous observation. We can do that by performing logistic regression (given by Eq. \eqref{eq:logisticregression}) in computing the parameter $\kappa$. To do that we use Pólya-Gamma augmentation (see Eq. \eqref{eq:polya_gamma}), which allows us for efficient Gibbs sampling. Therefore, we propose following model using notation that is in line with the conventions outlined in Section \ref{section:hdp}(presented in the Figure \ref{fig:pgm:rshdphmm}):

\begin{align}
\text { Transition prior: } & \kappa_{j, 1} \stackrel{i i d}{\sim} \operatorname{beta}\left(\rho_1, \rho_2\right), \\
& v_{j, t}=R_j^T y_t +r_j, \\
& \kappa_{j,t+1} = \frac{\left(e^{v_{j, t}}\right)}{1+e^{v_{j, t}}}, \\
& \bar{\pi}_j \stackrel{i i d}{\sim} \operatorname{DP}(\alpha \beta), \\
& \pi_{j,t}=\kappa_{j,t} \delta_j+\left(1-\kappa_{j,t}\right) \bar{\pi}_j, \\
& j=1,2, \ldots \text{ and } t=1, \ldots, T.\notag
\end{align}

An equivalent formulation of $z_t \sim \pi_{z_{t-1}}$ in the disentangled sticky HDP-HMM is as follows:
\begin{align}
\begin{array}{ll}
\text { Latent states : } & w_t \sim \operatorname{Bern}\left(\kappa_{z_{t-1}, t}\right), \\
& z_t \sim w_t \delta_{z_{t-1}}+\left(1-w_t\right) \bar{\pi}_{z_{t-1}}, \\
& t=1, \cdots, T,\notag
\end{array}
\end{align}
In this way, our self-persistance probability becomes position- and state-dependent. When considering motion, the likelihood of transitioning to a different state is contingent upon one's location.
\section{Inference and Learning}
\begin{figure}[htb]

\begin{algorithm}[H]
    \caption{Direct assignment sampler for RS-HDP-HMM}
    \label{alg:directassignment}
    \begin{algorithmic}[1]
\STATE Sequentially sample $\left\{z_t, w_t, w_{t+1}\right\}$ for $t=1, \ldots, T$.
\STATE Sample $\left\{\kappa_{j, 1}\right\}$ for $j=1, \ldots, K+1$, and compute $\left\{\kappa_{j, t+1}\right\}$ for $t=1, \ldots, T-1, j=1, \ldots, K+1$ $K$ is defined as number of unique states in $\left\{z_t\right\}_{t=1}^T$.
\STATE Sequentially sample auxiliary variables $\{\eta_{j, t}\}$ for $t=1, \ldots, T-1, j=1, \ldots, K+1$.
\STATE Sample $\beta$. {\tiny(Same as HDP-HMM).}
\STATE Sample hyperparameter $\alpha, \gamma, \rho_1, \rho_2, R$.
\end{algorithmic}
\end{algorithm}

\end{figure}
\subsection{Direct Assignment Sampler}

The direct assignment sampler marginalizes transition distributions $\pi_j$ and parameters $\theta_j$ and sequentially samples $z_t$ given all the other states $z_{\backslash t}$, observations $\left\{y_t\right\}_{t=1}^T$, and the global transition distribution $\beta$.

Our direct assignment sampler is based on the direct assignment sampler for DS-HDP-HMM, which means that instead of only sampling $z_t$ (as in the sampler for HDP-HMM), we sample $\left\{z_t, w_t, w_{t+1}\right\}$ in blocks.
We sample $\alpha, \beta, \gamma$ only using $z_t$ that switch to other states by $\bar{\pi}_{z_{t-1}}$ $\left(w_t=0\right)$, and sample $\left\{\kappa_{j, t}\right\}_{j=1}^{K+1}, \rho_1, \rho_2$ only using $z_t$ that stick to state $z_{t-1}$ $\left(w_t=1\right)$. The main difference between the samplers presented by \citet{fox_bayesian_2011, hutter_disentangled_2021} comes from the fact that $\kappa_{j}$ is now a vector and we have to additionally sample the auxiliary variables.

Algorithm \ref{alg:directassignment} presents direct assignment sampler steps. For Step 1, we compute the probability of each possible posterior case of $p\left(z_t, w_t, w_{t+1} \mid z_{\backslash t}, w_{\backslash\{t, t+1\}},\left\{y_t\right\}_{t=1}^T, \alpha, \beta,\left\{\kappa_{j, 1:T}\right\}_{j=1}^{K+1}\right)$ in sequence and sample the corresponding categorical distribution for $\left\{z_t, w_t, w_{t+1}\right\}$, where all states except $z_t$ are denoted by $z_{\backslash t}$, and all $w_t$ except for $w_t$ and $w_{t+1}$ are represented by $w_{\backslash\{t, t+1\}}$.
If $z_t=K+1$, that is, a new state appears, we increment $K$, sample the regression parameters $R_{K+1}$ for the new state from the prior state, and update $\beta$ using stick breaking. This requires $\mathcal{O}(T K)$ operations.

For Step 2, given $w_{t+1}$ whose corresponding $z_t$ is $j$, we sample $\kappa_{j,1}$ using beta-binomial conjugacy and compute $\kappa_{j, t}$ using Equation \eqref{eq:logisticregression}, performing the $\mathcal{O}(T K)$ steps.

Step 3 involves sampling $T(K+1)$ auxiliary variables, requiring
$\mathcal{O}(T K)$ steps.

Step 4 involves introducing auxiliary variables $\left\{m_{j k}\right\}_{j, k=1}^K$ and sampling $\beta$ using Dirichlet categorical conjugacy, which requires $\mathcal{O}(K)$ draws.

Step 5 computes the empirical transition matrix $\left\{n_{j k}\right\}_{j, k=1}^K$, where $n_{j k}$ is the number of transitions from state $j$ to $k$ with $w_t=0$ in $\left\{z_t\right\}_{t=1}^T$, and introduces additional auxiliary variables.
Then the posteriors of $\alpha$ and $\gamma$ are conjugated with Gamma, given the auxiliary variables. We approximate the posterior of $\rho_1, \rho_2$ by finite grids. This last step has a computational complexity of $\mathcal{O}(K)$. The total complexity per iteration is $\mathcal{O}(T K)$.

As mentioned in the study by \citet{fox_bayesian_2011}, when the entire latent sequence $\left\{z_t, w_t\right\}_{t=1}^T$ is sampled jointly, it significantly improves the mixing rate. This is particularly crucial for models with dynamics, such as ARHMM and SLDS, where the correlated observations can further slow down the mixing rate of the direct assignment sampler. For this reason, we did not utilize this sampler in our experiments. However, we have included its description for the sake of comprehensiveness in this study.

\subsection{Weak-Limit Sampler}

The weak-limit sampler for the sticky HDP-HMM takes advantage of the fact that the Dirichlet process is a discrete measure to produce a finite approximation of the HDP prior. This approximation converges to the HDP prior when the number of components, $L$, goes to infinity\citep{ishwaran_markov_2000,ishwaran_dirichlet_2002}. The standard HMM forward-backward procedure can be used to sample latent variables $\left\{z_t\right\}_{t=1}^T$ with the help of this approximation, which increases the mixing rate of the Gibbs sampler. Our weak-limit Gibbs sampler is based on the sampler for DS-HDP-HMM, so it samples pairs $\left\{z_t, w_t\right\}_{t=1}^T$.

\begin{figure}[htb]

\begin{algorithm}[H]
    \caption{Weak-limit sampler for RS-HDP-HMM}
    \label{alg:weak}
    \begin{algorithmic}[1]
\STATE Jointly sample $\left\{z_t, w_t\right\}_{t=1}^T$.
\STATE Sample $\left\{\kappa_{j, 1}\right\}$ for $j=1, \ldots, L$, and compute $\left\{\kappa_{j, t+1}\right\}$ for $t=1, \cdots, T-1, j=1, \ldots, L$.
\STATE Sequentially sample auxiliary variables $\{\eta_{j, t}\}$ for $t=1, \ldots, T-1, j=1, \ldots, L$.
\STATE Sample $\left\{\beta_j\right\}_{j=1}^L,\left\{\bar{\pi}_j\right\}_{j=1}^L$. {\tiny(Same as HDP-HMM).}
\STATE Sample $\left\{\theta_j\right\}_{j=1}^L$. 
\STATE Sample hyperparameter $\alpha, \gamma, \rho_1, \rho_2, R$.
\end{algorithmic}
\end{algorithm}

\end{figure}

Algorithm \ref{alg:weak} presents weak-limit sampler steps. In Step 1, we use the forward-backward procedure to jointly sample the two-dimensional latent variables $\left\{z_t, w_t\right\}_{t=1}^T$.

Step 2 is the same as in the direct assignment sampler.

Step 3 is the same as in the direct assignment sampler.

For Step 4, we sample $\beta$ and $\bar{\pi}$ based on Dirichlet-categorical conjugacy, given the auxiliary variables $\left\{m_{j k}\right\}_{j, k=1}^L$, the empirical transition matrix $\left\{n_{j k}\right\}_{j, k=1}^L$, and the approximate prior.

Step 5 involves placing a conjugate prior in $\theta_j$ and using conjugacy to sample the posterior. Step 5 is the same as in the direct assignment sampler. The complexity of each step is $\mathcal{O}\left(T L^2\right), \mathcal{O}(L)$, $\mathcal{O}(L), \mathcal{O}(L)$, and $\mathcal{O}(L)$, respectively, with a total complexity of $\mathcal{O}\left(T L^2\right)$.


The beam sampling technique \citep{van_gael_beam_2008, dewar_inference_2012, slupinski_improving_2024} is a useful method for sampling the approximation parameter L, eliminating the need to pre-set it while still allowing the transition matrix to be instantiated and the state sequence to be block sampled. This is especially beneficial when the number of states is too large to be accurately estimated beforehand. In this paper, we will not examine beam sampling.

\section{Experiments}
\begin{figure}[!h]
\centering
\includegraphics[width=0.75\linewidth]{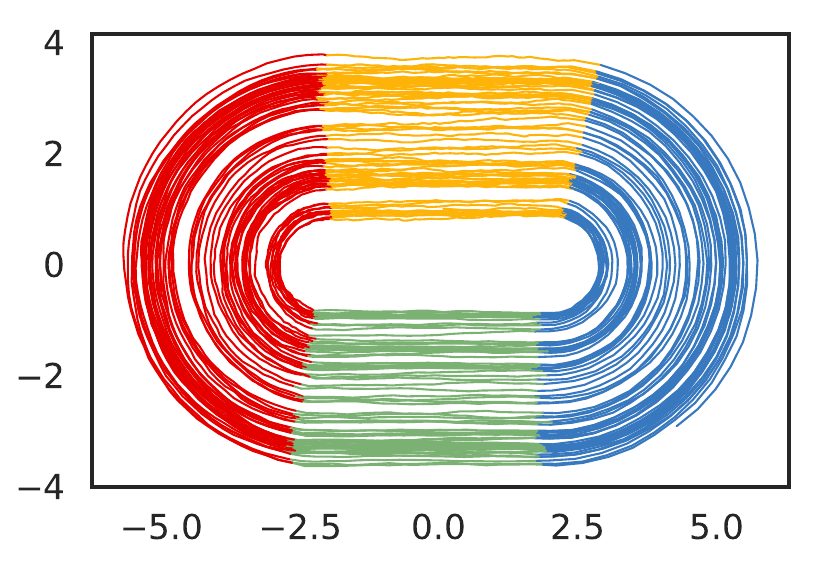}
\caption{Trajectory of \emph{NASCAR $^{\circledR}$} used to test the models.}
\label{fig:nascar:trajectory}
\end{figure}
To evaluate our motion segmentation model's effectiveness, we compared it against HDP-HMM, sticky HDP-HMM, and disentangled sticky HDP-HMM on three benchmarks. The HDP-HMM benchmarks used the implementation from \citet{hutter_disentangled_2021}.\footnote{\url{https://github.com/zhd96/ds-hdp-hmm}}

We employed the Matrix Normal inverse-Wishart ($\operatorname{MNIW}(M, V, S, n)$) prior for autoregressive emission in all experiments, detailed in \citet{fox_bayesian_2011}. Specifically, $M_0=\mathbf{0}$ and $V_0=I$ were selected to center the prior's mass around stable matrices. The inverse-Wishart component had $n_0=m+2$ degrees of freedom, with the scale matrix $S_0$ set to $0.4\bar{\Sigma}$, where $\bar{\Sigma}$ is defined as $\frac{1}{T-1} \sum\left(x_t-\overline{x}\right)\left(x_t-\overline{x}\right)^T$ and $x_t = y_{t+1} - y_{t}$. 

All models used a Gamma (1, 0.01) prior for concentration parameters $\alpha$, and $\alpha+\kappa$ for sticky HDP-HMM. To avoid numerical instability in sampling $\beta$ by preventing extreme $\gamma$ values, a Gamma (2,1) prior was used for $\gamma$. 

Self-persistence parameters employed non-informative priors: a uniform distribution in [0, 1] for $\phi=\frac{\rho_1}{\rho_1+\rho_2}$, and for DS-HDP-HMM, a uniform distribution in [0, 2] for $\eta=\left(\rho_1+\rho_2\right)^{-1 / 3}$. The support of $\phi$ and $\eta$ was partitioned into a $100 \times 100$ grid for simulated data. The prior for $(R, r)$ was $\mathcal{N}(0, 0.0001\cdot I)$.

Log-likelihood was our primary metric to compare models, with higher values indicating a better fit. This metric provides a straightforward probabilistic comparison, as employed in \citet{hutter_disentangled_2021, fox_bayesian_2009}. We also used accuracy and F1 scores for assessing segmentation quality.

For quantitative evaluation, we used the dancing bee and \emph{NASCAR®} benchmarks, widely used in Bayesian motion segmentation studies \citep{oh_learning_2008, fox_bayesian_2011, linderman_bayesian_2017, nassar_tree-structured_2019}. For qualitative evaluation, we chose the mouse behavior dataset, a key dataset in self-supervised segmentation research \citep{hutter_disentangled_2021,batty_behavenet_2019,costacurta_distinguishing_2022}.
All of the sampler runs were performed on Intel Xeon Platinum 8268 using two cores.

\subsection{NASCAR $^{\circledR}$}

We begin with a straightforward illustration where the dynamics take the form of oval shapes, resembling a stock car on a \emph{NASCAR®} track. The dynamics is determined by four distinct states, $z_t \in\{1, \ldots, 4\}$, which regulate a continuous latent state in two dimensions, $y_t \in \mathbb{R}^2$.

For this experiment, we performed ten runs with different random seeds. The results are averaged.

\begin{figure}[!h]

\centering
\includegraphics[width=0.8\linewidth]{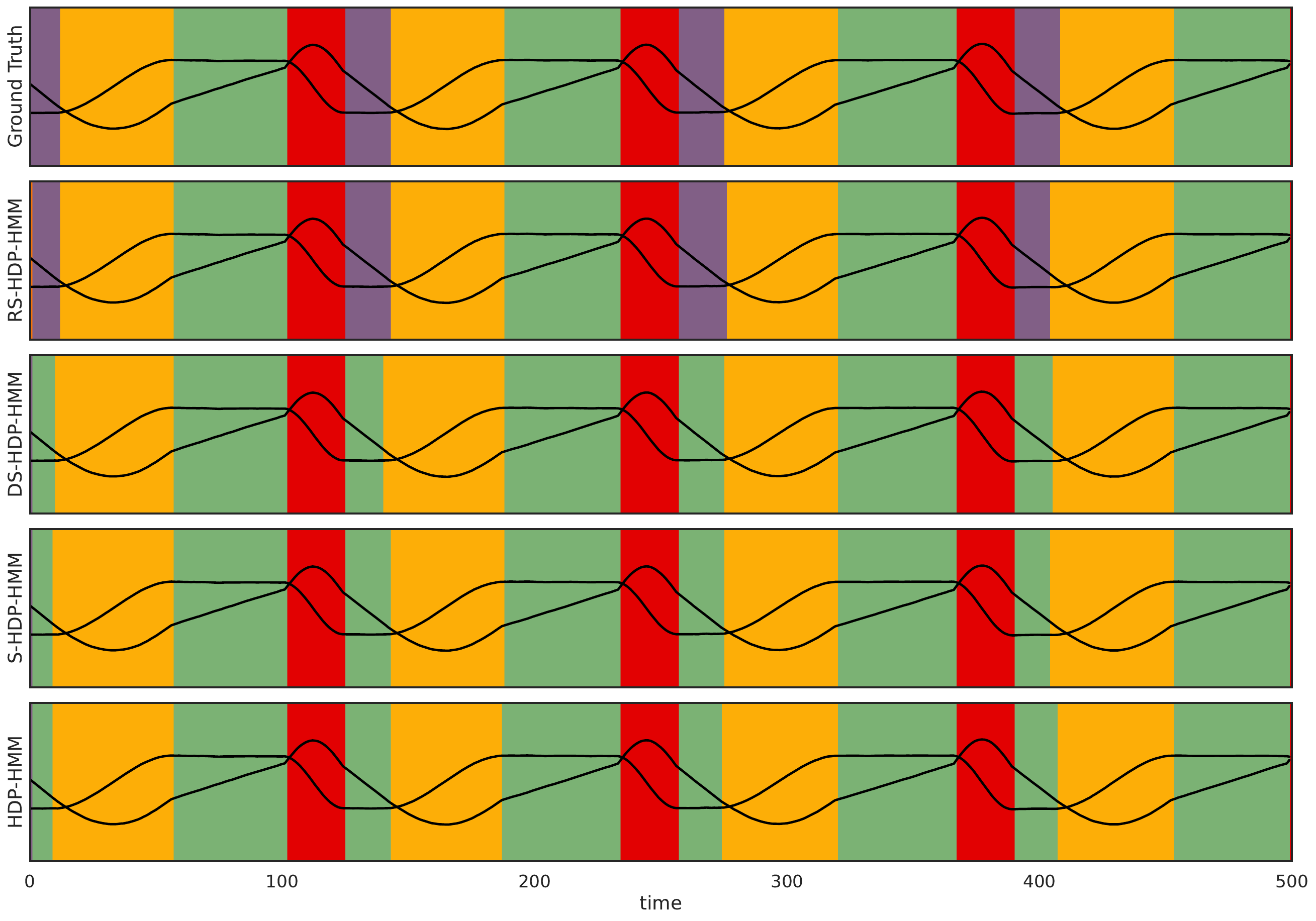}
\caption{Sample results of \emph{NASCAR $^{\circledR}$} segmentation. The color represents an assigned segment. Methods other than RS-HDP-HMM often fail to differentiate two "straight" states.}
\label{fig:nascar_seg}
\end{figure}

We can distinguish four states here, which are presented in Figure \ref{fig:nascar:trajectory}. To make the states more distinguishable, the car uses four different velocities (which can be observed in Figure \ref{fig:nascar_seg}, as the segments differ significantly in length).

It is evident that the state of the system is influenced by its position. As the car completes each lap, it spends a greater amount of time in the "straight-on" states. This poses a challenge for models that have a constant transition matrix, as the probability of self-persistence follows a Geometric distribution.
\begin{figure}[!h]

\subfigure[HDP-HMM\label{fig:nascar_models_ll:hdp}]{
  \centering
  \includegraphics[width=0.45\linewidth]{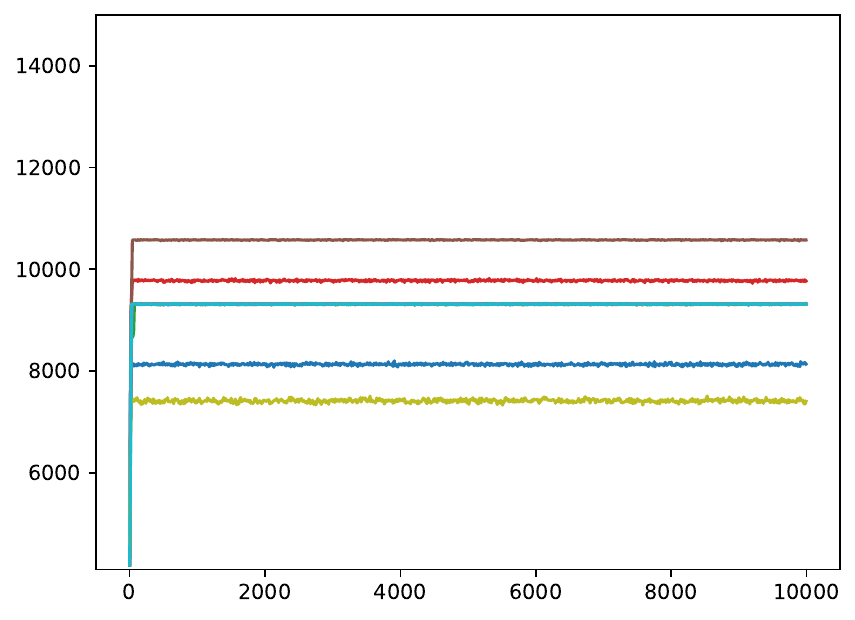}
}
  \subfigure[S-HDP-HMM\label{fig:nascar_models_ll:shdp}]{
  \centering
  \includegraphics[width=0.45\linewidth]{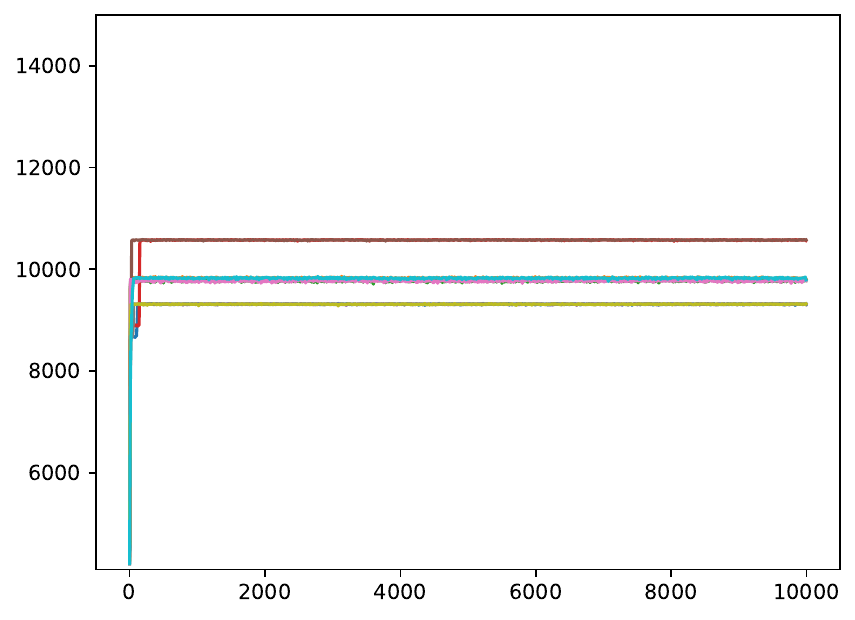}
}
\newline
  \subfigure[DS-HDP-HMM\label{fig:nascar_models_ll:rdshdp}]{
  \centering
  \includegraphics[width=0.45\linewidth]{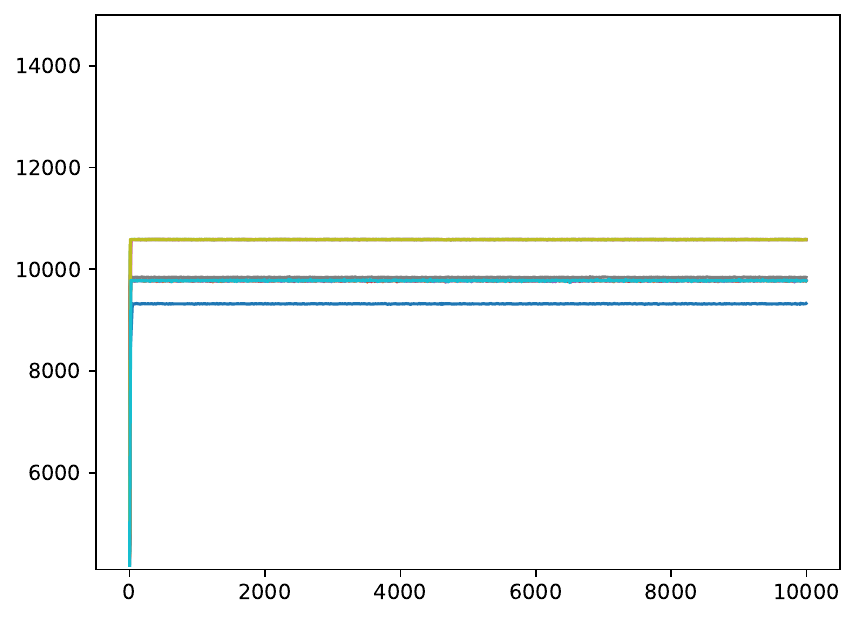}
}%
\subfigure[RS-HDP-HMM\label{fig:nascar_models_ll:rdshdp}]{
  \centering
  \includegraphics[width=0.45\linewidth]{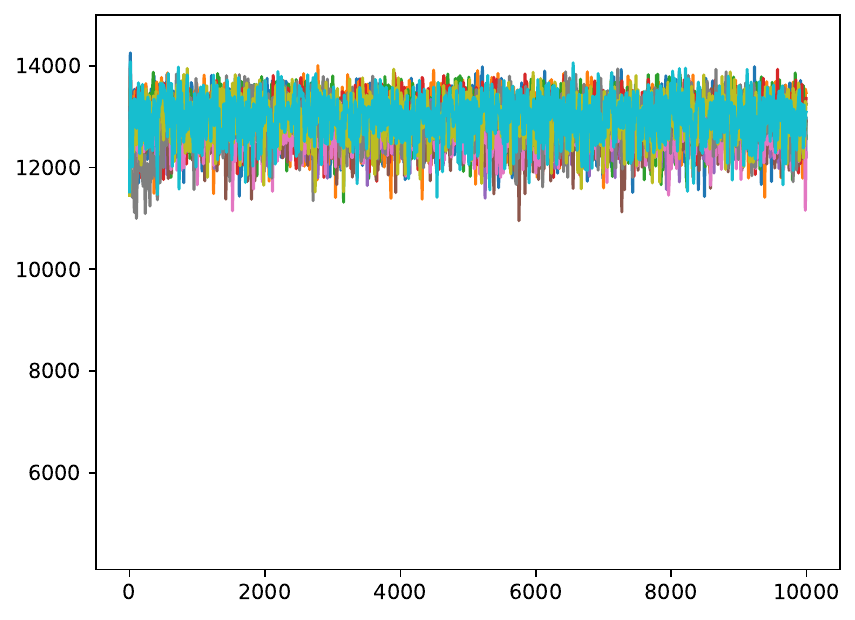}
}%
\caption{Log-likelihood during models' fitting on \emph{NASCAR $^{\circledR}$} dataset. Our model obtains the highest likelihood scores. It presents a diverse range of likelihood scores within a single run. Additionally, we observe that the other models often converge to local optima. Furthermore, the RS-HDP-HMM model exhibits a considerably shorter burn-in time.}%
\label{fig:nascar_models_ll}
\end{figure}

\begin{table}[!htb]
\centering
\caption{Results of \emph{NASCAR $^{\circledR}$} data segmentation.}
\label{tab:nascar:results}
\begin{tabular}{lrr}
\toprule
{} &  Weighted F1 &  Accuracy \\
\midrule
HDP-HMM         &         0.86 &      0.82 \\
S-HDP-HMM               &         0.91 &      0.88 \\
DS-HDP-HMM               &         0.94 &      0.95 \\
RS-HDP-HMM  &        \textbf{0.97} &      \textbf{0.97} \\
\bottomrule
\end{tabular}
\end{table}

The segmentation outcomes are demonstrated in Table \ref{tab:nascar:results}. It is evident that our model exhibits superior performance. Figure \ref{fig:nascar_seg} illustrates that the remaining models occasionally struggle to differentiate the fourth state.

\begin{figure}[!h]
\centering
\includegraphics[width=0.8\linewidth]{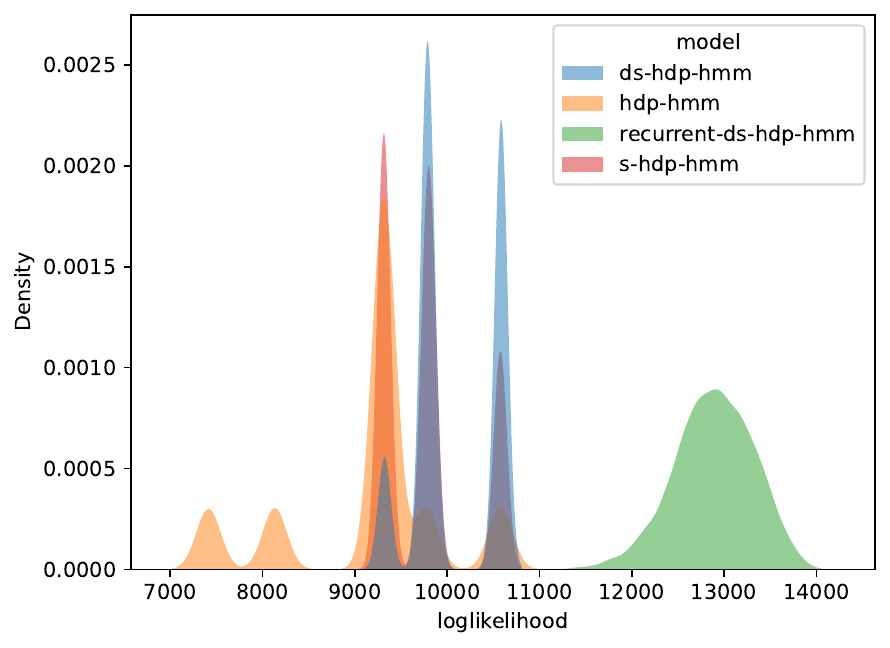}
\caption{Distribution of \emph{NASCAR$^{\circledR}$} log-likelihoods taken after 200 burn-in iterations.
Our model attains the highest likelihood scores and generates the most coherent results. However, it is worth noting that it also demonstrates the greatest variability in likelihood within a single execution.
}
\label{fig:nascar:ll_dist}
\end{figure}

The results obtained from Figures \ref{fig:nascar:ll_dist} and \ref{fig:nascar_models_ll} demonstrate that our model achieves the highest likelihood scores and produces the most coherent outcomes. Nevertheless, it is important to mention that it also exhibits the largest variance in likelihood in a single run.


\subsection{Dancing Bee}
We employed the publicly available dancing bees dataset \citep{oh_learning_2008}, which is renowned for its intricate characteristics and has previously been studied in the field of time series segmentation.

The dataset consists of the trajectories followed by six honey bees while performing the waggle dance, including their 2D coordinates and heading angle at each time step. The bees demonstrate three types of motion: waggle, turn right, and turn left.

\begin{table}[!htb]
\centering
\caption{Results of dancing bee data segmentation.}
\label{tab:bee:results}
\begin{tabular}{lrr}
\toprule
{} &  Weighted F1 &  Accuracy \\
\midrule
HDP-HMM         &         0.73 &      0.77 \\
S-HDP-HMM               &         \textbf{0.85} &      \textbf{0.87} \\
DS-HDP-HMM               &         0.83 &      0.84 \\
RS-HDP-HMM  &         \textbf{0.85} &     \textbf{ 0.87} \\
\bottomrule
\end{tabular}
\end{table}

\begin{figure}[!htb]

\centering
\includegraphics[width=0.9\linewidth]{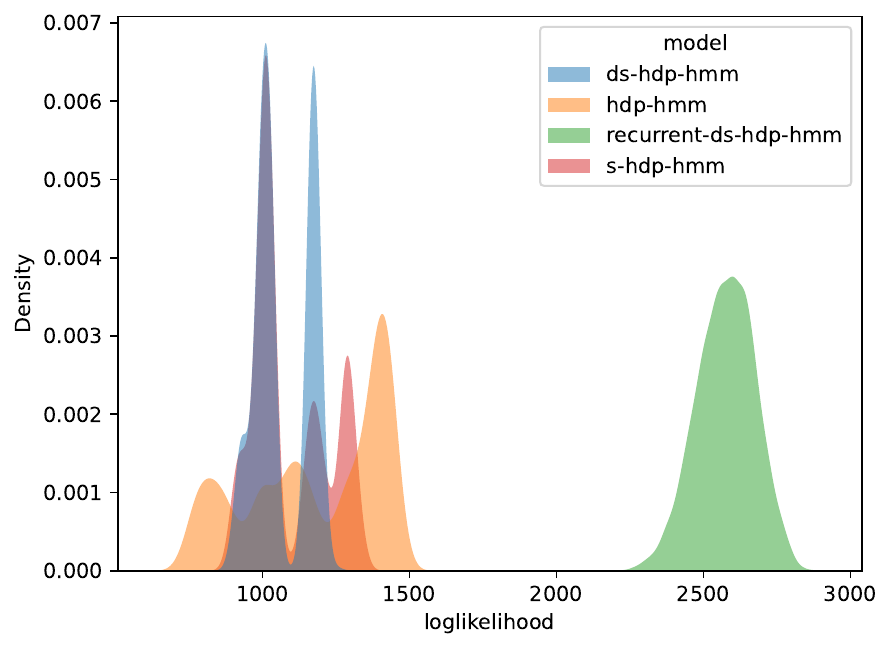}
\caption{Distribution of models' log-likelihoods taken after 2000 burn-in iterations on dancing bee dataset. Our model achieves the highest likelihood scores, demonstrating a clear unimodal distribution of likelihood scores. Furthermore, we observe that the other models often become trapped in local optima.}
\label{fig:bee:ll_dist}
\end{figure}

\begin{figure}[!htb]

\centering
\includegraphics[width=0.8\linewidth]{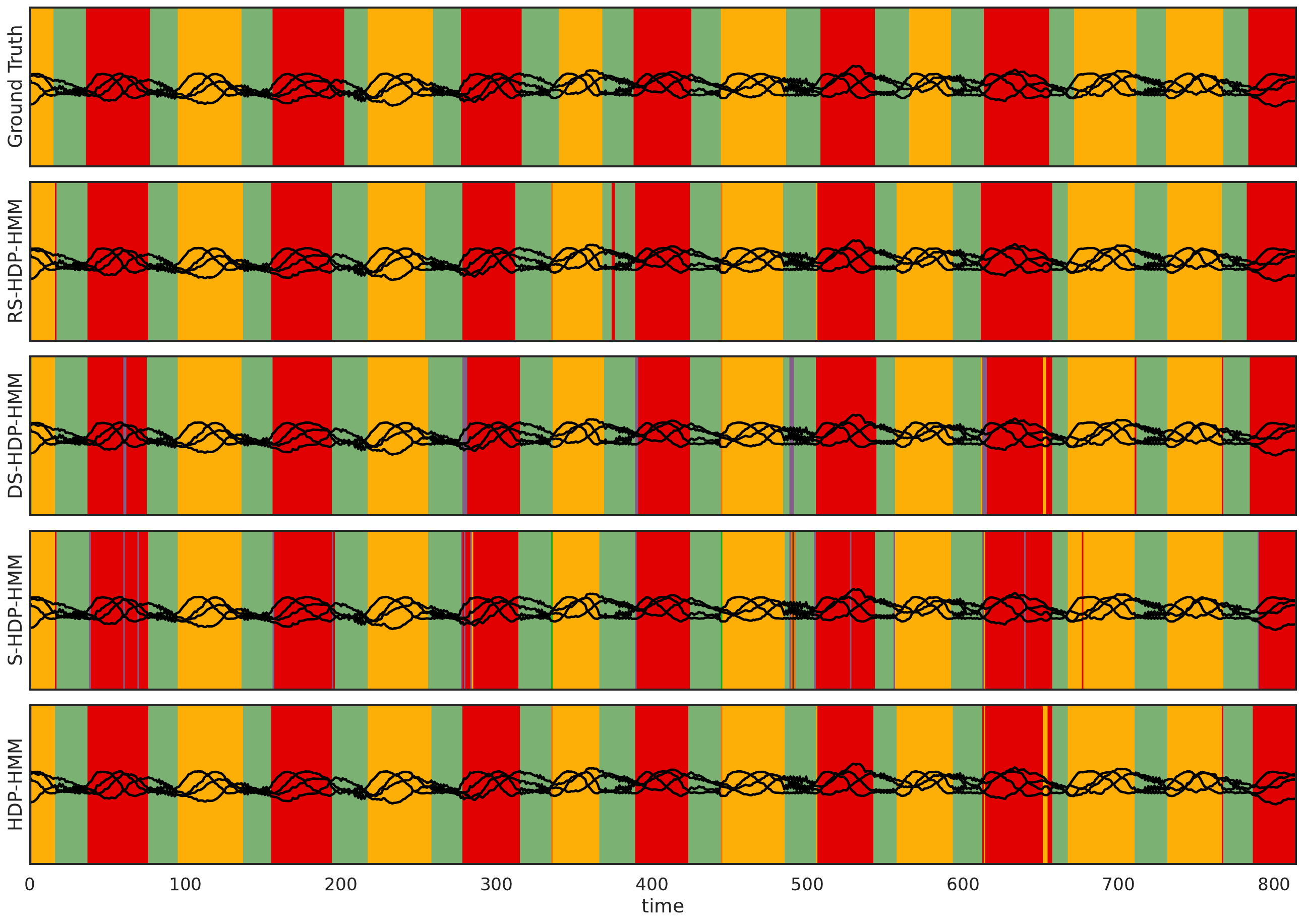}
\caption{Results of dancing bee dataset. The color represents an assigned segment. The segmentation obtained from our model is the smoothest.}
\label{fig:bee_seg}
\end{figure}

Honey bees communicate the whereabouts of food sources by performing a series of dances within the hive. These dances consist of waggle, turn right, and turn left movements. The waggle dance of a bee consists of it walking in a straight line while shaking its body from side to side quickly. The turning dances are simply the bee rotating clockwise or counterclockwise. The data consist of $\boldsymbol{y}_t=\left[\begin{array}{llll}\cos \left(\theta_t\right), & \sin \left(\theta_t\right), & x_t, & y_t\end{array}\right]^T$, where $\left(x_t, y_t\right)$ denotes the $2 \mathrm{D}$ coordinates of the body of the bee and $\theta_t$ its head angle.

For this experiment, we performed ten runs with different random seeds. The results are averaged.

We can see in Table \ref{tab:bee:results} that our model performed on a par with S-HDP-HMM in terms of accuracy and weighted F1. However, in Figure \ref{fig:bee_seg} we can observe that the segmentation obtained from our model is the smoothest.

In the same graph, it is evident that the values observed during the "waggle" phase are significantly smaller in magnitude compared to those obtained during turning. We hypothesize that this characteristic enables our model to achieve a more seamless segmentation.

\begin{figure}[!htb]
\subfigure[HDP-HMM\label{fig:bee_models_ll:hdp}]{
  \centering
  \includegraphics[width=0.45\linewidth]{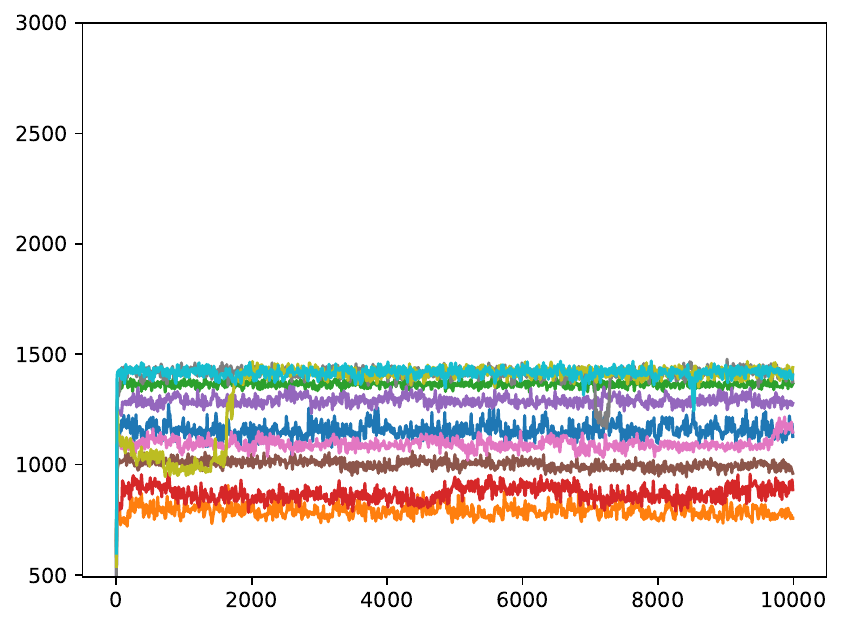}
}
  \subfigure[S-HDP-HMM\label{fig:bee_models_ll:shdp}]{
  \centering
  \includegraphics[width=0.45\linewidth]{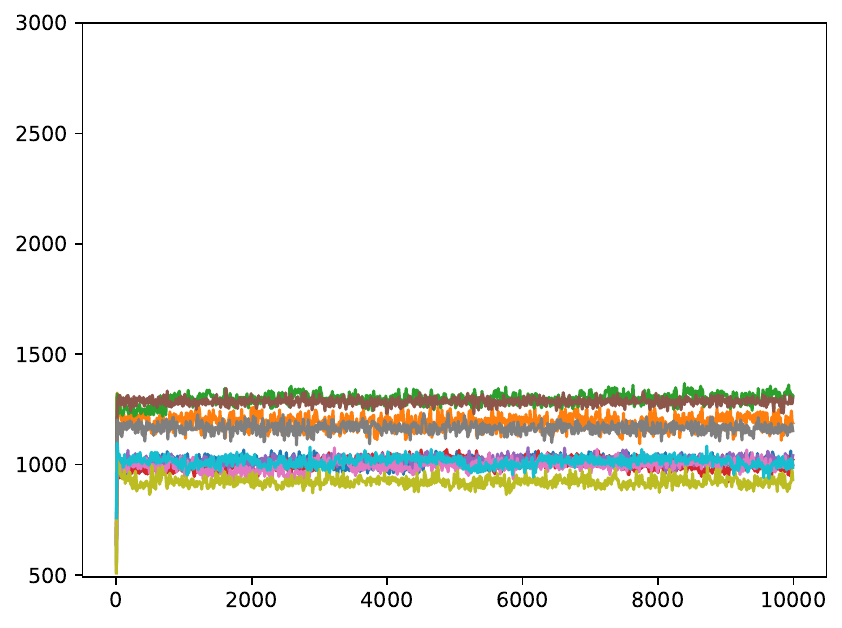}
}
\newline
  \subfigure[DS-HDP-HMM\label{fig:bee_models_ll:rdshdp}]{
  \centering
  \includegraphics[width=0.45\linewidth]{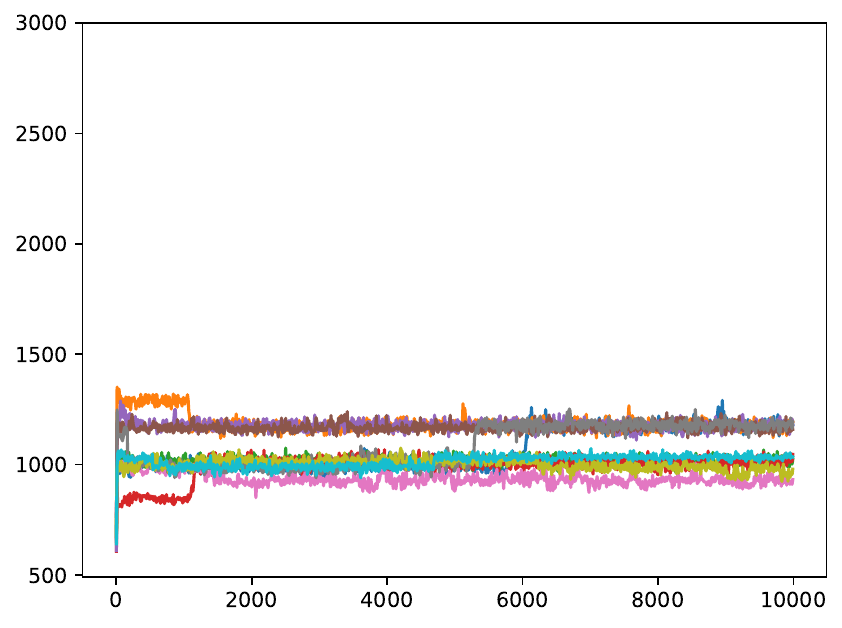}
}%
\subfigure[RS-HDP-HMM\label{fig:bee_models_ll:rdshdp}]{
  \centering
  \includegraphics[width=0.45\linewidth]{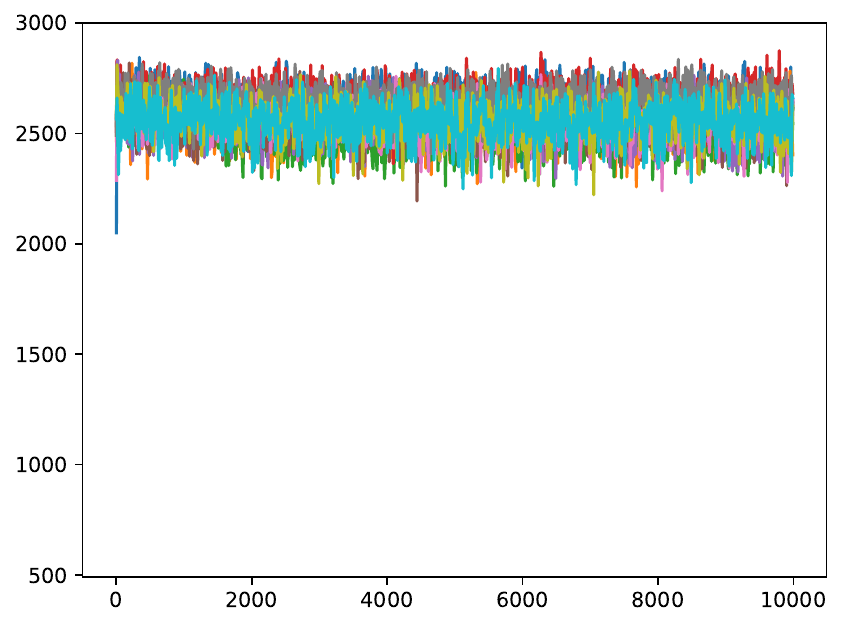}
}%
\caption{Log-likelihood during models' fitting on honey bee movements. Our model achieves the highest likelihood scores and generates the most coherent results. However, it demonstrates the widest range of likelihood scores within a single execution. We can also see that it has signficantly smaller burn-in time.}%
\label{fig:bee_models_ll}
\end{figure}

Like the previous experiment, the findings shown in Figures \ref{fig:bee:ll_dist} and \ref{fig:bee_models_ll} demonstrate that our model achieves the highest likelihood scores and produces the most coherent outcomes. Additionally, it exhibits the greatest range of likelihood scores in a single execution.

\subsection{Mouse Behavior}

Our model was also tested on a publicly accessible mouse behavior dataset. In the described experiment, the mouse was immobilized by fixing its head position during a visual decision-making task. During this task, neural activities in the dorsal cortex were monitored through wide-field calcium imaging. The behavioral video data comprised grayscale frames with a resolution of 128x128 pixels. This video was recorded from two different angles using two cameras, one for the side view and another for the bottom view. To manage the high-dimensional nature of the video data, we adopted dimension reduction techniques previously established. These involved 9-dimensional continuous variables derived through a convolutional autoencoder \citep{batty_behavenet_2019}.


\begin{figure}[!htb]

\centering
\includegraphics[width=0.8\linewidth]{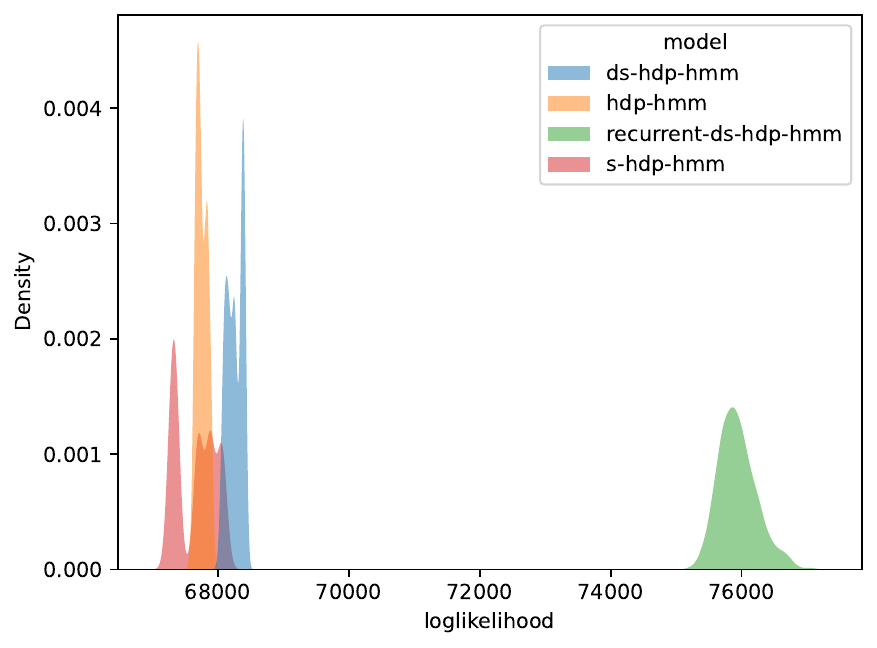}
\caption{Distribution of log-likelihoods on mouse behavior dataset taken after 9000 burn-in iterations. In line with previous findings, our model attains the highest likelihood scores, indicating a distinct unimodal distribution of likelihood scores with the greatest variation within each run. Additionally, we note that the other models do not converge to a shared optimal solution.}
\label{fig:behavenet:ll_dist}

\end{figure}

\begin{figure}[!htb]

\centering
\includegraphics[width=0.8\linewidth]{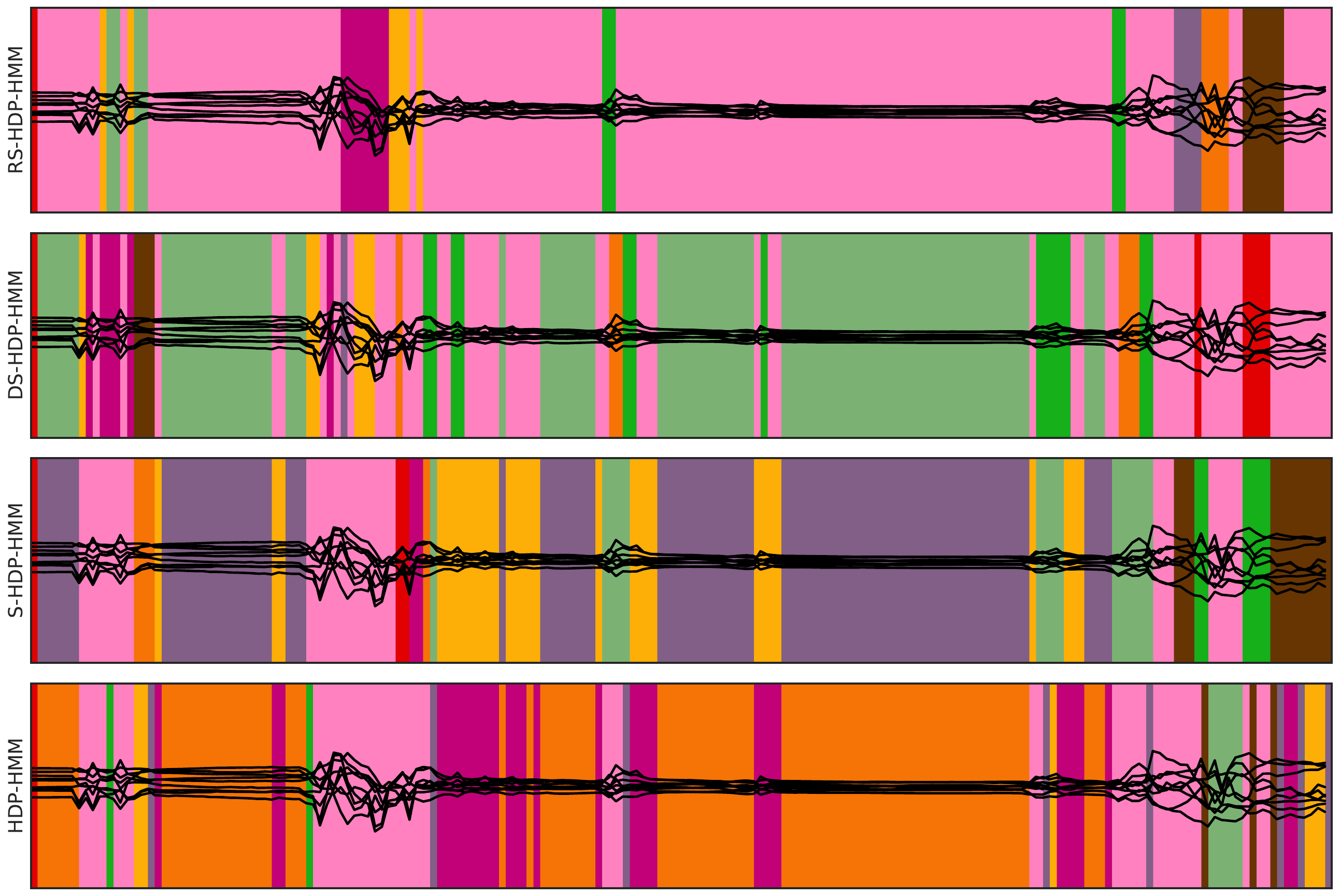}
\caption{Results of mouse behavior dataset. The color represents an assigned segment. One issue frequently encountered in ARHMM models is known as over-segmentation. However, RS-HDP-HMM segmentation, despite utilizing a similar number of states as DS-HDP-HMM, offers a smoother and more stable alternative.}
\label{fig:behavenet_seg}

\end{figure}


In real-world scenarios, ARHMM models often suffer from a problem called over-segmentation. This refers to the situation where behaviors that seem similar to a human expert are divided into separate clusters. According to the hypothesis proposed by \citet{costacurta_distinguishing_2022}, over-segmentation occurs because the ARHMM model confuses discrete factors of behavioral variability, such as the expression of different behavioral syllables, with continuous factors of variability that cannot be adequately represented using linear autoregressive dynamics with Gaussian noise.

We performed five independent MCMC runs for all model types. On average, RS-HDP-HMM converged to 33 states, DS-HDP-HMM to 34, S-HDP-HMM to 21 and HDP-HMM to 23 states.
On average there were 6143 state switches for RS-HDP-HMM,  7713 for DS-HDP-HMM, 6412 for S-HDP-HMM and 7801 for HDP-HMM.

Despite using nearly as many states as DS-HDP-HMM, RS-HDP-HMM segmentation remains much smoother and more stable. A sample segmentation is presented in Figure \ref{fig:behavenet_seg}.
For example, in the pink state, our model demonstrates a greater tolerance for minor noise. This is because it considers not only the direction of position change but also the actual position. Therefore, even if the direction changes, as long as the position change is small enough, there will be no transition to a different state.


\section{Conclusions and Future Work}

The paper presents a new modeling technique that combines latent states with position- and state-dependent self-persistence probability. This technique is particularly suitable for motion data, where the likelihood of transitioning to a different state depends on the subject's location.

A central contribution of this paper is introducing dependence of stickiness factor on the previous observations.

The RS-HDP-HMM addresses a significant limitation in conventional HMMs: their inability to effectively model non-Markovian temporal dependencies. Many real-world processes exhibit behaviors where the current state depends not just on the immediate past state (as in hidden Markov models), but on a more complex history of previous states. Similar level of contribution was introduced by RSLDS or iRSLDS bringing similar dependece to SLDS.

The recurrent model consistently demonstrates better performance metrics, showcasing its strength in capturing data nuances, and the obtained states are more robust to small noise, as it is shown by mouse behavior dataset.

Similarly to \cite{hutter_disentangled_2021} who modified the algorithms introduced by \cite{fox_bayesian_2009}, we introduce two sampling schemes for our model.


\section*{Acknowledgements}
Calculations have been carried out using resources provided by Wroclaw Centre for Networking and Supercomputing (http://wcss.pl), grant No. 405. 

This work was supported by the Polish National Science Centre (NCN) under grant OPUS-18 no. 2019/35/B/ST6/04379.
\bibliography{bibliography}

\appendix
\onecolumn
\section{Sampling Scheme}

Step 1 is analogous to \citet{fox_bayesian_2011} and \citet{hutter_disentangled_2021}. We describe it here for completeness.
\paragraph{Step 1: Sequentially Sample $\boldsymbol{z}_{\boldsymbol{t}}, \boldsymbol{w}_{\boldsymbol{t}}, \boldsymbol{w}_{\boldsymbol{t}+\boldsymbol{1}}$} : The posterior distribution of $z_t$, $w_t$, and $w_{t+1}$ is expressed as follows:
\begin{equation}
\begin{aligned}
p\left(z_t\right. & \left.=k, w_t, w_{t+1} \mid z_{\backslash t}, w_{\backslash\{t, t+1\}}, y_{1: T}, \alpha, \beta,\left\{\kappa_{j, 1:T}\right\}_{j=1}^{K+1}\right) \\
\propto & p\left(z_t=k, w_t, w_{t+1} \mid z_{\backslash t}, w_{\backslash\{t, t+1\}}, \alpha, \beta,\left\{\kappa_{j, 1:T}\right\}_{j=1}^{K+1}\right) \cdot p\left(y_t \mid y_{\backslash t}, z_t=k, z_{\backslash t}\right) .
\end{aligned}    
\end{equation}

where all observations except $y_t$ are denoted by $y_{\backslash t}$, and all $w_t$ except for $w_t$ and $w_{t+1}$ are represented by $w_{\backslash\{t, t+1\}}$.

The predictive observation likelihood $p\left(y_t \mid y_{\backslash t}, z_t=k, z_{\backslash t}\right)$ can be quickly calculated if we use a conjugate prior on the parameter in the observation likelihood.
\begin{equation}\label{eq:suppl_posterior}
    \begin{aligned}
& p\left(z_t=k, w_t, w_{t+1} \mid z_{\backslash t}, w_{\backslash\{t, t+1\}}, \alpha, \beta,\left\{\kappa_{j,1:T}\right\}_{j=1}^{K+1}\right) \\
\propto & p\left(z_t=k, w_t, w_{t+1}, z_{t+1} \mid z_{\backslash\{t, t+1\}}, w_{\backslash\{t, t+1\}}, \alpha, \beta,\left\{\kappa_{j, 1:T}\right\}_{j=1}^{K+1}\right) \\
\propto & \int_{\bar{\pi}} p\left(w_t \mid \kappa_{z_{t-1}, t}\right) p\left(z_t \mid w_t, \bar{\pi}_{z_{t-1}}\right) p\left(w_{t+1} \mid \kappa_{z_t, t+1}\right) p\left(z_{t+1} \mid w_{t+1}, \bar{\pi}_{z_t}\right) \\
& \prod_i\left(\prod_{\tau \mid z_{\tau-1}=i, w_\tau=0, \tau \neq t, t+1} p\left(\bar{\pi}_{i} \mid \alpha, \beta\right) p\left(z_\tau \mid \bar{\pi}_{i}\right)\right) d \bar{\pi} \\
\propto & \int_{\bar{\pi}} p\left(w_t \mid \kappa_{z_{t-1}, t}\right) p\left(z_t \mid w_t, \bar{\pi}_{z_{t-1}}\right) p\left(w_{t+1} \mid \kappa_{z_t, t+1}\right) p\left(z_{t+1} \mid w_{t+1}, \bar{\pi}_{z_t}\right) \\
& \prod_i p\left(\bar{\pi}_{i} \mid\left\{\tau \mid z_{\tau-1}=i, w_\tau=0, \tau \neq t, t+1\right\}, \alpha, \beta\right) d \bar{\pi} .
\end{aligned}
\end{equation}

Let $z_{t-1}=j, z_{t+1}=l$, then Equation \eqref{eq:suppl_posterior} simplifies to
\newpage
\scalebox{0.9}{$
\begin{cases}
\kappa_{j, t}\kappa_{j, t+1}, & \text { if } w_t=w_{t+1}=1, k=j=l \\ 
\left(1-\kappa_{j, t}\right) \kappa_{l, t+1} \int_{\bar{\pi}_{j}} p\left(z_t=l \mid w_t, \bar{\pi}_j\right) p\left(\bar{\pi}_{j} \mid\left\{\tau \mid z_{\tau-1}=j, w_\tau=0, \tau \neq t\right\}, \alpha, \beta\right) d \bar{\pi}_{j}, & \text { if } w_t=0, w_{t+1}=1, k=l \\
\left(1-\kappa_{j, t}\right) \kappa_{j, t} \int_{\bar{\pi}_{j}} p\left(z_{t+1}=l \mid w_{t+1}, \bar{\pi}_{j}\right) p\left(\bar{\pi}_{j} \mid\left\{\tau \mid z_{\tau-1}=j, w_\tau=0, \tau \neq t+1\right\}, \alpha, \beta\right) d \bar{\pi}_{j}, & \text { if } w_t=1, w_{t+1}=0, k=j \\ 
\left(1-\kappa_{j, t}\right)\left(1-\kappa_{k, t+1}\right) \int_{\bar{\pi}_{j}} p\left(z_t=k \mid w_t, \bar{\pi}_{j}\right) p\left(\bar{\pi}_{j} \mid\left\{\tau \mid z_{\tau-1}=j, w_\tau=0, \tau \neq t\right\}, \alpha, \beta\right) d \bar{\pi}_{j} . & \\
\int_{\bar{\pi}_{k}} p\left(z_{t+1}=l \mid w_{t+1}, \bar{\pi}_{k}\right) p\left(\pi_{k, t+1} \mid\left\{\tau \mid z_{\tau-1}=k, w_\tau=0, \tau \neq t+1\right\}, \alpha, \beta\right) d \bar{\pi}_{k}, & \text { if } w_t=0, w_{t+1}=0, k \neq j \\
\left(1-\kappa_{j, t}\right)\left(1-\kappa_{k, t+1}\right) & \\
\int_{\bar{\pi}_{j}} p\left(z_t=j \mid w_t, \bar{\pi}_{j,t}\right) p\left(z_{t+1}=l \mid w_{t+1}, \bar{\pi}_{j}\right) p\left(\pi_j \mid\left\{\tau \mid z_{\tau-1}=j, w_\tau=0, \tau \neq t, t+1\right\}, \alpha, \beta\right) d \bar{\pi}_j, & \text { if } w_t=0, w_{t+1}=0, k=j \\
0, & \text { otherwise }\end{cases}
$}

The result is the following equations:
\begin{equation}\label{eq:suppl_frac}
\begin{aligned}
& \int_{\bar{\pi}_j} p\left(z_t=k \mid w_t, \bar{\pi}_j\right) p\left(\bar{\pi}_j \mid\left\{\tau \mid z_{\tau-1}=j, w_\tau=0, \tau \neq t\right\}, \alpha, \beta\right) d \bar{\pi}_j=\frac{\alpha \beta_k+n_{j k}^{-t}}{\alpha+n_{j \cdot}^{-t}}, \\
& \int_{\bar{\pi}_k} p\left(z_{t+1}=l \mid w_{t+1}, \bar{\pi}_k\right) p\left(\bar{\pi}_k \mid\left\{\tau \mid z_{\tau-1}=k, w_\tau=0, \tau \neq t+1\right\}, \alpha, \beta\right) d \bar{\pi}_k=\frac{\alpha \beta_l+n_{k l}^{-t}}{\alpha+n_{k .}^{-t}} \\
& \int_{\bar{\pi}_j} p\left(z_t=j \mid w_t, \bar{\pi}_j\right) p\left(z_{t+1}=l \mid w_{t+1}, \bar{\pi}_j\right) p\left(\bar{\pi}_j \mid\left\{\tau \mid z_{\tau-1}=j, w_\tau=0, \tau \neq t, t+1\right\}, \alpha, \beta\right) d \bar{\pi}_j \\
& =\frac{\left(\alpha \beta_j+n_{j j}^{-t}\right)\left(\alpha \beta_l+n_{j l}^{-t}+\delta(j, l)\right)}{\left(\alpha+n_{j .}^{-t}\right)\left(\alpha+n_{j \cdot}^{-t}+1\right)}, \\
& \int_{\bar{\pi}_j} p\left(z_t=k \mid w_t, \bar{\pi}_j\right) p\left(\bar{\pi}_j \mid\left\{\tau \mid z_{\tau-1}=j, w_\tau=0, \tau \neq t\right\}, \alpha, \beta\right) d \bar{\pi}_j . \\
& \int_{\bar{\pi}_k} p\left(z_{t+1}=l \mid w_{t+1}, \bar{\pi}_k\right) p\left(\bar{\pi}_k \mid\left\{\tau \mid z_{\tau-1}=k, w_\tau=0, \tau \neq t+1\right\}, \alpha, \beta\right) d \bar{\pi}_k \\
& =\frac{\alpha \beta_k+n_{j k}^{-t}}{\alpha+n_{j .}^{-t}} \frac{\alpha \beta_l+n_{k l}^{-t}}{\alpha+n_{k .}^{-t}}
\end{aligned}
\end{equation}

where the quantity $n_{z_{t-1} z_t}$ is the number of times that the transition from state $z_{t-1}$ to state $z_t$ occurred with $w_t=0$ in the sequence $z_{1: T}$. Furthermore, $n_{z_{t-1} z_t}^{-t}$ is the number of times the transition from state $z_{t-1}$ to $z_t$ occurred with $w_t=0$, not including the transitions from $z_{t-1}$ to $z_t$ or from $z_t$ to $z_{t+1}$. Lastly, $n_j$ is the total number of transitions from state $j$ to any other state.

If $z_t$ is equal to $K+1$, that is, a new state has appeared, we will increase $K$, draw a probability of self-persistence $\kappa_{K+1}$ for the new state from the prior, and update $\beta$ using the stick-breaking method. We will sample $b$ from a $\operatorname{beta}(1, \gamma)$ distribution, and assign $\beta_K=b \beta_{k_{\text {new }}}, \beta_{k_{\text {new }}}=$ $(1-b) \beta_{k_{\text {new }}}$, where $\beta_{k_{\text {new }}}=\sum_{i=K+1}^{\infty} \beta_i$.

\paragraph{Step 2: Sample $\left\{\kappa_{j, 1}\right\}$ for $j=1, \cdots, K+1$, and compute $\left\{\kappa_{j, t+1}\right\}$}
The posterior distribution of $\kappa_{j, 1}$ is obtained by applying the beta-binomial conjugate property. The expression for the posterior distribution is given by: 
\begin{equation}
    \kappa_{j, 1} \sim \operatorname{beta}\left(\rho_1+\sum_{\tau, z_{\tau-1}=j} w_\tau, \rho_2+\sum_{\tau, z_{\tau-1}=j} 1-w_\tau\right), j=1, \cdots, K+1
\end{equation}
Here, $\kappa_{K+1}$ represents the self-persistence probability of a new state.

For $t+1$ $\left\{\kappa_{j, t+1}\right\}$ can be computed:
\begin{equation}
\begin{gathered}
\kappa_{j, t+1}=\frac{\left(e^{v_{j, t}}\right)}{1+e^{v_{j, t}}} \\
v_{j, t}=R_j^T x_t+r_j
\end{gathered}    
\end{equation}

\paragraph{Step 3: Sequentially sample the auxiliary variables $\{\eta_{j, t}\}$ for $t=1, \cdots, T-1, j=1, \cdots, L$.}
7. The conditional posteriors of the Pólya-Gamma random variables are also Pólya-Gamma: $\eta_{j, t} \mid z_t,\left(R_j, r_j\right), x_{t-1} \sim \mathrm{PG}\left(1, \nu_{j, t}\right)$.

We used samplers implemented in \url{https://github.com/zoj613/polyagamma}.
\paragraph{Step 4: Sample $\boldsymbol{\beta}$} We introduce auxiliary variables $m_{j k}$ to sample the global transition distribution $\beta$. This is done using the Chinese restaurant franchise (CRF) formulation of HDP prior \cite{teh_hierarchical_2006}. These variables can be thought of as the number of tables in restaurant $j$ serving dishes $k$. We first update $m_{j k}$ and then sample $\beta$.

For each $(j, k) \in\{1, \cdots, K\}^2$, set $m_{j k}$ and $s$ to zero. Then, for $i=1, \cdots, n_{j k}$, draw a sample from a Bernoulli distribution with probability $\frac{\alpha \beta_k}{n+\alpha \beta_k}$. Increase $s$ and if the sample is equal to one, increment $m_{j k}$. We can then sample $\beta$ as 
\begin{equation}
    \left(\beta_1, \beta_2, \cdots, \beta_K, \beta_{k_{\text {new }}}\right) \sim \operatorname{Dir}\left(m_{\cdot 1}, \cdots, m_{\cdot K}, \gamma\right), 
\end{equation}
where $\beta_{k_{\text {new }}}=\sum_{i=K+1}^{\infty} \beta_i$ is for transitioning to a new state, and $m_{\cdot k}$ is $\sum_{j=1}^K m_{j k}$. A more thorough explanation of the sampling of $\beta$ can be found in \cite{fox_bayesian_2011, hutter_disentangled_2021}.

\paragraph{Step 5: Sample Hyperparameters $\alpha, \gamma, \rho_1, \rho_2, R, r$} We use the same values of $\alpha$ and $\gamma$ as in \cite{teh_hierarchical_2006, escobar_bayesian_1995}. By introducing some extra variables, we can make the posterior of $\alpha$ and $\gamma$ follow a gamma distribution if we assign a gamma prior to them. To handle $\rho_1, \rho_2$, we use the reparametrization technique discussed in Chapter 5 of \cite{rubin_bayesian_2015}. This involves transforming $\phi=\frac{\rho_1}{\rho_1+\rho_2}, \eta=\left(\rho_1+\rho_2\right)^{-1 / 3}$, and assigning a Uniform $([0,1] \times[0,2])$ prior on $(\phi, \eta)$. Then, we can discretize the support of $(\phi, \eta)$ and calculate the posterior numerically.

Let $\eta_{j, t}$ be the auxiliary Pólya-gamma random variable introduced at time $t$. We can sample values of $R_j, r_j$ from the following posterior:
\begin{equation}
    p\left(\left(R_j, r_j\right) \mid x_{1: T}, z_{1: T}, \eta_{j, 1: T}\right) \propto p\left(\left(R_j, r_j\right)\right) \prod_{t=1}^T \mathcal{N}\left(v_{j, t} \mid \lambda_{j, t} / \eta_{j, t}, 1 / \eta_{j, t}\right)^{\mathbbm{1}\left(z_t=j\right)},
\end{equation}
where $\lambda_{j, t}=w_{t+1} - \frac{1}{2}$. 
\section{Weak limit-sampler}

\paragraph{Step 1: Sample $\left\{z_t, w_t\right\}_{t=1}^T$} The combined conditional distribution of $z_{1: T}, w_{1: T}$ is determined by
\begin{equation}
    \begin{aligned}
p\left(z_{1: T}, w_{1: T} \mid y_{1: T}, \bar{\pi},\left\{\kappa_j\right\}_{j=1}^L, \theta\right)= & p\left(z_T, w_T \mid z_{T-1}, y_{1: T}, \bar{\pi},\left\{\kappa_{j, 1:T}\right\}_{j=1}^L, \theta\right) \\
& p\left(z_{T-1}, w_{T-1} \mid z_{T-2}, y_{1: T}, \bar{\pi},\left\{\kappa_{j, 1:T}\right\}_{j=1}^L, \theta\right) \\
& \cdots p\left(z_1 \mid y_{1: T}, \bar{\pi},\left\{\kappa_{j,1:T}\right\}_{j=1}^L, \theta\right) .
\end{aligned}
\end{equation}

The conditional probability distribution of $z_1$ is given by:
\begin{equation}
    p\left(z_1 \mid y_{1: T}, \bar{\pi},\left\{\kappa_{j,1:T}\right\}_{j=1}^L, \theta\right) \propto p\left(z_1\right) p\left(y_1 \mid \theta_{z_1}\right) p\left(y_{2: T} \mid z_1, \bar{\pi},\left\{\kappa_{j,2:T}\right\}_{j=1}^L, \theta\right)
\end{equation}

The conditional distribution of $z_t, t>1$ is given by:
\begin{equation}
\begin{aligned}
& p\left(z_t, w_t \mid z_{t-1}, y_{1: T}, \bar{\pi},\left\{\kappa_{j,1:T}\right\}_{j=1}^L, \theta\right) \\
\propto & p\left(z_t, w_t, y_{1: T} \mid z_{t-1}, \bar{\pi},\left\{\kappa_{j,1:T}\right\}_{j=1}^L, \theta\right) \\
= & p\left(z_t \mid \bar{\pi}_{z_{t-1}}, w_t\right) p\left(w_t \mid z_{t-1},\left\{\kappa_{j,t}\right\}_{j=1}^L\right) p\left(y_{t: T} \mid z_t, \bar{\pi},\left\{\kappa_{j, t:T}\right\}_{j=1}^L, \theta\right) p\left(y_{1: t-1} \mid z_{t-1}, \bar{\pi},\left\{\kappa_{j, 1:t-1}\right\}_{j=1}^L, \theta\right) \\
\propto & p\left(z_t \mid \bar{\pi}_{z_{t-1}}, w_t\right) p\left(w_t \mid z_{t-1},\left\{\kappa_{j,t}\right\}_{j=1}^L\right) p\left(y_{t: T} \mid z_t, \bar{\pi},\left\{\kappa_{j,t: T}\right\}_{j=1}^L, \theta\right) \\
= & p\left(z_t \mid \bar{\pi}_{z_{t-1}}, w_t\right) p\left(w_t \mid z_{t-1},\left\{\kappa_{j,t}\right\}_{j=1}^L\right) p\left(y_t \mid \theta_{z_t}\right) m_{t+1, t}\left(z_t\right),
\end{aligned}
\end{equation}

The backward message passed from $z_t$ to $z_{t-1}$, denoted by $m_{t+1, t}\left(z_t\right)$, is the probability of observing $y_{t+1: T}$ given $z_t$, $\bar{\pi}$, $\left\{\kappa_{j,t}\right\}_{j=1}^L$, and $\theta$. This is recursively defined as follows:
\begin{equation}
\begin{aligned}
& m_{t+1, t}\left(z_t\right)=\sum_{z_{t+1}, w_{t+1}} p\left(z_{t+1} \mid \bar{\pi}_{z_t}, w_{t+1}\right) p\left(w_{t+1} \mid z_t,\left\{\kappa_{j, t+1}\right\}_{j=1}^L\right) p\left(y_{t+1} \mid \theta_{z_{t+1}}\right) m_{t+2, t+1}\left(z_{t+1}\right), t \leq T \\
& m_{T+1, T}\left(z_T\right)=1
\end{aligned}
\end{equation}

\paragraph{Step 2: Sample $\left\{\kappa_{j, 1}\right\}$ for $j=1, \cdots, K+1$, and compute $\left\{\kappa_{j, t+1}\right\}$} Same as in direct assignment sampler step 2, but for $j=1, \cdots, L$.

\paragraph{Step 3: Sample $\left\{\beta_j\right\}_{j=1}^L,\left\{\bar{\pi}_j\right\}_{j=1}^L$}
\begin{equation}
\begin{aligned}
\beta \mid m, \gamma & \sim \operatorname{Dir}\left(\gamma / L+m_{\cdot 1}, \cdots, \gamma / L+m_{\cdot L}\right), \\
\bar{\pi}_j \mid z_{1: T}, w_{1: T}, \alpha, \beta & \sim \operatorname{Dir}\left(\alpha \beta_1+n_{j 1}, \cdots, \alpha \beta_L+n_{j L}\right), j=1, \cdots, L .
\end{aligned}
\end{equation}

\paragraph{Step 4: Sample $\left\{\boldsymbol{\theta}_j\right\}_{j=1}^L$} We draw $\theta$ from the posterior distribution based on the emission function and the base measure $H$ of the parameter space $\Theta$, i.e. $\theta_j \mid z_{1: T}, y_{1: T} \sim p\left(\theta_j \mid\left\{y_t \mid z_t=j\right\}\right)$. For all the emissions in the main paper, we can conveniently sample $\theta_j$ from its posterior due to the conjugacy properties.

\paragraph{Step 5: Sample Hyperparameters $\alpha, \gamma, \rho_1, \rho_2$} Same as step 4 in direct assignment sampler.

\section{MNIW Prior}

In the autoregressive emission, the MNIW prior is established by assigning a matrix-normal prior $\mathcal{M N}(M, \Sigma_j, V)$ to $A_j$ conditioned on $\Sigma_j$:
\begin{equation}
    p\left(A_j \mid \Sigma_j\right)=\frac{1}{(2 \pi)^{d^2 / 2}|V|^{d / 2}\left|\Sigma_j\right|^{d / 2}} \exp \left(-\frac{1}{2} \operatorname{tr}\left[\left(A_j-M\right)^{\top} \Sigma_j^{-1}\left(A_j-M\right) V^{-1}\right]\right),
\end{equation}
where $M$ is $d \times d$ matrix, $\Sigma_j, V$ are $d \times d$ positive-definite matrix, $\mathrm{d}$ is the dimension of observation $y_t$; and an inverse-Wishart prior $I W\left(S_0, n_0\right)$ on $\Sigma_j$ :
\begin{equation}
    p\left(\Sigma_j\right)=\frac{\left|S_0\right|^{n_0 / 2}}{2^{n_0 d / 2} \Gamma_d\left(n_0 / 2\right)}\left|\Sigma_j\right|^{-\left(n_0+d+1\right) / 2} \exp \left(-\frac{1}{2} t r\left(\Sigma_j^{-1} S_0\right)\right),
\end{equation}
where \(\Gamma_d(\cdot)\) denotes the multivariate gamma function. We define \(M=\mathbf{0}\), \(V=I_{d \times d}\), \(n_0=d+2\), and \(S_0=0.4 \bar{\Sigma}\), with \(\bar{\Sigma}\) being \(\frac{1}{T} \sum_{t=1}^{T-1}(x_t-\bar{x})(x_t-\bar{x})^\top\), and \(x_t = y_{t+1} - y_t\).

\end{document}